\documentclass[journal]{IEEEtran}

\usepackage{fullpage, amsmath, amsfonts, amssymb, amsthm, amstext}
\usepackage{enumitem}
\usepackage[numbib]{tocbibind}
\usepackage{graphicx}
\usepackage{subfigure}
\usepackage{stmaryrd}
\usepackage{color}
\usepackage{mdframed}
\usepackage{multirow}
\DeclareGraphicsExtensions{.pdf,.png,.jpg}
\usepackage[margin=1in]{geometry}
\usepackage{nccmath}
\usepackage{tikz-cd}
\usepackage{listings}
\usepackage{mathrsfs}
\usepackage{cite}

                    {$\blacksquare$\vspace*{7pt}} 

\def\R{{\mathbb R}}

\def\bi{{\mathbf i}}

\def\w{{\boldsymbol w}}

\def\x{\boldsymbol{x}}
\def\y{{\boldsymbol y}}
\def\z{{\boldsymbol z}}

\def\W{{\mathbf W}}

\def\bi{\begin{itemize}} \def\ei{\end{itemize}}
\def\be{\begin{eqnarray*}}
\def\ee{\end{eqnarray*}}

\def\etal{{\it et al }}

\def\0{{\mathbf 0}}

\newcommand{\beq}{\begin{equation}}
\newcommand{\eeq}{\end{equation}}

\def\XXint#1#2#3{{\setbox0=\hbox{$#1{#2#3}{\int}$ }
\vcenter{\hbox{$#2#3$ }}\kern-.55\wd0}}

\begin{document}
%
\title{Improving learnability of neural networks:
 adding supplementary axes to disentangle data representation}
%
%

\author{  Bukweon~Kim,
Sung~Min~Lee$^*$
		and~Jin~Keun~Seo
		\thanks{B. Kim, S.M. Lee, and J. K. Seo are with the Department of Computational Science and Engineering,
			Yonsei University, Seoul 03722, South Korea. Asterisk indicates the corresponding author (e-mail:sungminlee@yonsei.ac.kr).}
	}

%



\maketitle

\begin{abstract}
Over-parameterized deep neural networks have proven to be able to learn an arbitrary dataset with 100$\%$ training accuracy. Because of a risk of overfitting and computational cost issues, we cannot afford to increase the number of network nodes if we want achieve better training results for medical images.
Previous deep learning research shows that the training ability of a neural network improves dramatically (for the same epoch of training) when a few nodes with supplementary information are added to the network. These few informative nodes  allow the network to learn features that are otherwise difficult to learn by generating a disentangled data representation.
This paper analyzes how concatenation of additional information as supplementary axes affects the training of the neural networks. This analysis was conducted for a simple multilayer perceptron (MLP) classification model with a rectified linear unit (ReLU) on two-dimensional training data.
We compared the networks with and without concatenation of supplementary information to support our analysis. The model with concatenation showed more robust and accurate training results compared to the model without concatenation. We also confirmed that our findings are valid for deeper convolutional neural networks (CNN) using ultrasound images and for a conditional generative adversarial network (cGAN) using the MNIST data.
\end{abstract}


\IEEEpeerreviewmaketitle

\section{Introduction}

\IEEEPARstart{D}{eep} learning approaches have achieved significant progress in medical imaging tasks such as segmentation, diagnostics, detection, generating data, and image reconstruction \cite{Dorj2018,Fraz2012,Hyun2018,Kim2018,Yi2018}.
Along with these achievements, there have been several theoretical studies on why deep learning is so successful \cite{Eigen2013,Pascanu2013,Glorot2010,Zeiler2014,Rahaman2018}.
One of the main issues in neural network research is a generalization gap, which is the difference between the training error and the test error.
It is well known that over-parameterized deep neural networks (i.e. deep neural networks which use significantly more parameters than the number of samples in the training data) can learn an arbitrary dataset with 100$\%$ training accuracy \cite{Zhang2017}. The over-parameterized networks have a risk of overfitting. However, previous research \cite{Rahaman2018} shows that deep neural networks tend to learn low frequencies of the target function first, which prevents them from serious overfitting.
In medical image field, being highly data hungry, we cannot simply increase the number of nodes to improve the trainability at the risk of overfitting.

Several deep learning approaches show that learning ability improved dramatically when a few nodes with contextual information about the data were added to the low-performing network.
In a deep learning study that required a semantic segmentation of ultrasound images \cite{Kim2018}, concatenating spine information into the network as global information improved the segmentation accuracy when it was hard to get the correct segmentation. A conditional generative adversarial network cGAN \cite{Mirza2014} generates a data distribution more similar to the actual data distribution than that of a generative adversarial network (GAN) \cite{Goodfellow2014} by adding information about the conditions to the generator and discriminator in GAN. The deep learning paper for undersampled MRI \cite{Hyun2018} shows that learnability is changed dramatically by adding only a single phase encoding line to an undersampled $k$-space data. All these improvements can be considered as adding supplementary axes to the data in the form of additional nodes of the network.
These axes ease the handling of data by disentangling the data representation in a distinct manner.

This paper analyzes how concatenation of additional information as supplementary axes in the network affects the training of the neural networks. For visual explanation, we focused on classifying data distributed in two-dimensional (2D) input domain using multilayer perceptron (MLP) with a rectified linear unit (ReLU). We use MLP structures to show how the input space is distinguished by ReLU in the output space. We observe the effect of adding supplementary axes to the data by comparing three models (model A, model B, model C): model A is the standard structure, model B is the model A with an added node containing prior information, and model C is the model A with a node added without any information.
The comparison showed that adding a node with prior information of data to the network facilitated linear separation by disentangling the data representation in a higher-dimensional space and allowed for high nonlinear representation with fewer nodes.

To support our results, we conducted experiments on a convolutional neural network (CNN) \cite{LeCun2015} that classifies the center point of an `image patch' from an ultrasound image. In this classification, a relative position of the image patch with respect to the spine position is used as additional information. In this experiment, we saw that additional information dramatically increases the performance of the target network compared to other similar networks. We also compared the result between GAN and cGAN with the Modified National Institute of Standards and Technology (MNIST) dataset. By using a condition vector (such as class label) as supplementary information, the cGAN generates a data distribution closer to the test data distribution compared to that of GAN.

\section{Methods}
In this section, we analyze how concatenation of additional information as supplementary nodes of a network affects the learnability of the neural network. For the experiment, we considered an MLP model which is a function $F:\Bbb R^{n_0}\rightarrow\Bbb R^{n_L}$ defined as
\begin{equation}
F(\x)=h_{L}\circ h_{L-1} \circ \cdots \circ h_{1}(\x),
\end{equation}
where $h_l=\sigma_l(\W_lh_{l-1}+\boldsymbol{b}_l)$ is a composition of non-linear activation function and linear activation function.
Here, $\sigma_l$ is considered as ReLU for $l\in\{1,2,\cdots,L-1\}$, $\sigma_L$ is a softmax function, $\W_l\in\Bbb R^{n_{l-1}\times n_{l}}$ are weight matrices, and $\boldsymbol{b}_l \in\Bbb R^{n_l}$ are bias vectors.
For the each layer of MLP with ReLU, when the ReLU changes its behavior at zero, the function $\Bbb R^{n_{l-1}} \rightarrow \Bbb R^{n_l}$; $\x_{l-1} \rightarrow \sigma_l(\W_l\x_{l-1}+\boldsymbol{b}_l)$ changes its behavior at all inputs from any of the hyperplanes $H_i:=\{\x_{l-1}\in\Bbb R^{n_{l-1}} : \w^{i}_l\x_{l-1}+b_l^i=0 \}$ for $i\in\{1,2,\cdots,n_{l}\}$.

\begin{figure}[h]
\centering
\includegraphics[scale=1.2]{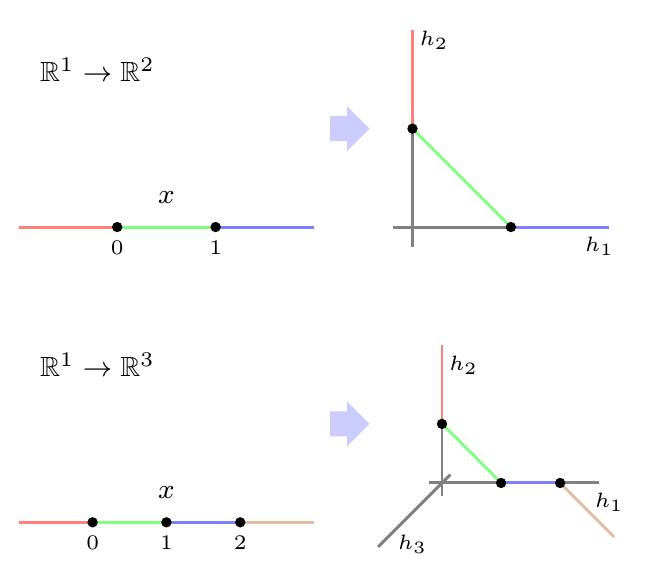}
\caption{The role of a single layer. The examples are for the cases of $\R^1\rightarrow\R^2$ and $\R^1\rightarrow\R^3$.}
\label{relu1}
\end{figure}

\begin{figure}[h]
\centering
\includegraphics[scale=1]{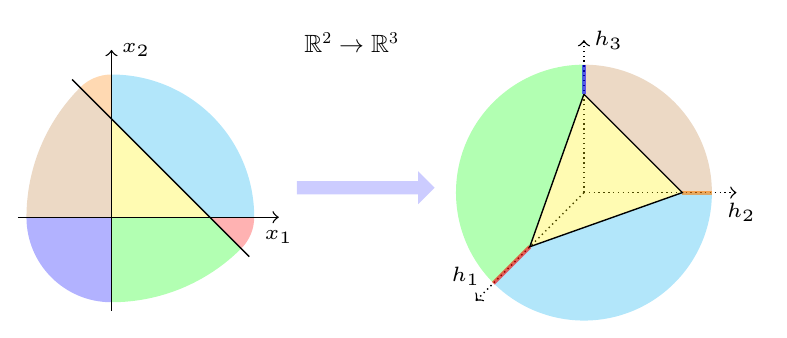}
\caption{The role of a single layer. The example is for the case of $\R^2\rightarrow\R^3$.}
\label{relu2}
\end{figure}

For example, as shown in Fig. \ref{relu1}, the function $\Bbb R^{1} \rightarrow \Bbb R^{2}$ maps a one-dimensional (1D) input data distribution into several linear regions diveded by the points which are 1D-sense hyperplanes. We can also obaserve that the function $\Bbb R^{1} \rightarrow \Bbb R^{3}$ with more nodes provides more distinguished linear regions. Similarly, in Fig. \ref{relu2} we distinguish 2D input data distribution into several linear regions divided by three lines which are 2D-sense hyperplanes.



\begin{figure*}[h]
\centering
\includegraphics[width=1\textwidth]{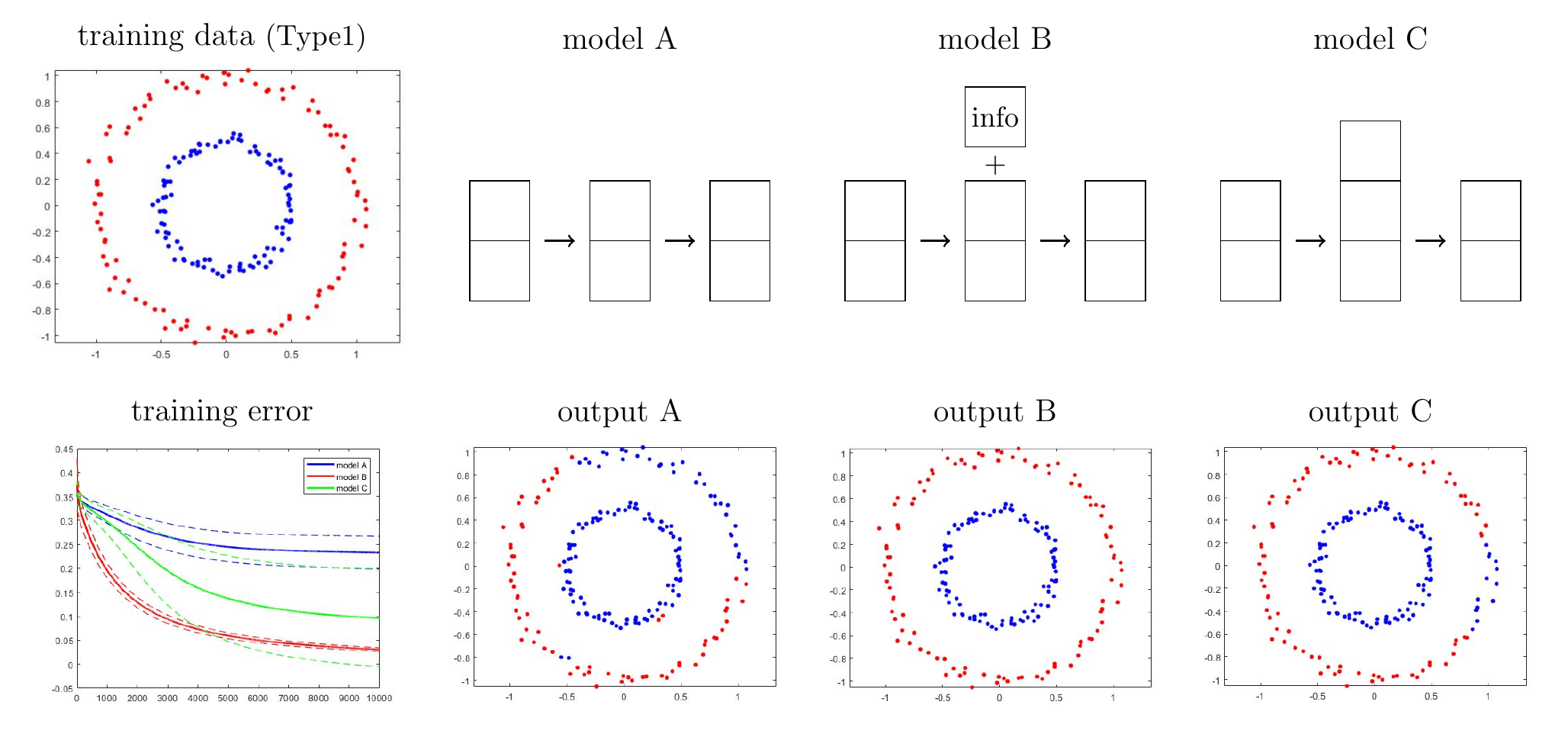}
\caption{Illustraion of the training data of `Type1' and three network models. Each model had been trained 1000 times with random initial weighs. Training error graph shows mean and standard deviation of the training errors for each model (solid line is mean and dashed line is standard deviation). The output is a training result of each model.}
\label{fig:toy1}
\end{figure*}

For simplicity, we focused on classifying data distributed on two-dimensional (2D) input domain using MLP with ReLU.
Let $(x_1,x_2)\in\Bbb R^2$ represent an input data.
We consider the training data of `Type1' and the networks, as shown in Fig. \ref{fig:toy1}. The training data of `Type1' was generated by adding a Gaussian noise to the data located at the distances of 0.5 and 1 from the center zero. To observe the effect of added information, we considered three models (model A, model B, model C); model A is the standard structure, model B is the model A with an added node containing the distance information ($\sqrt{{x_1}^2+{x_2}^2}$), and model C is the model A with a node added without any information.
Each model had been trained 1000 times with random initial weighs, and each training had 10000 epochs. The mean and standard deviation of the training errors for each model are shown in Fig. \ref{fig:toy1}.

\begin{figure}[h]
\centering
\includegraphics[width=.45\textwidth]{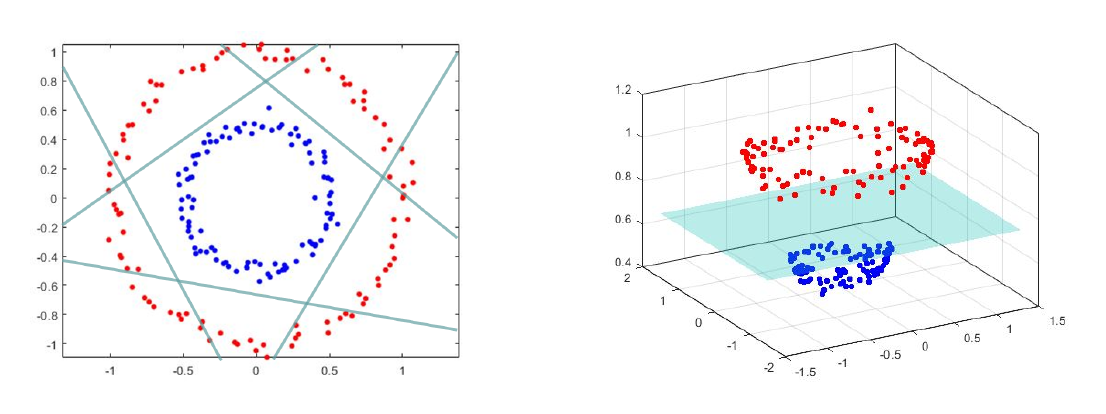}
\caption{Effect of data dimension on classification. A set of training data in a high-dimensional space is more likely to be linearly separable than in a low-dimensional space.}
\label{fig:toy1-1}
\end{figure}

In the training results, model A did not train at all but models B and C were trained. As a result, increasing the dimension of the feature map was helpful for training. In fact, according to Cover's theorem, a set of training data that is not linearly separable in a low-dimensional space is more likely to be linearly separable in a high-dimensional space. Specifically, as shown in Fig. \ref{fig:toy1-1}, we needed five linear hyperplanes to classify the `Type1' data in 2D space, but a good classification in 3D space can be achieved with one linear hyperplane. Furthermore, the training error graph on Fig. \ref{fig:toy1} shows that model B was trained significantly better than model C. That is, model C heavily depended on the initial values.

\begin{figure*}[h]
\centering
\includegraphics[width=1\textwidth]{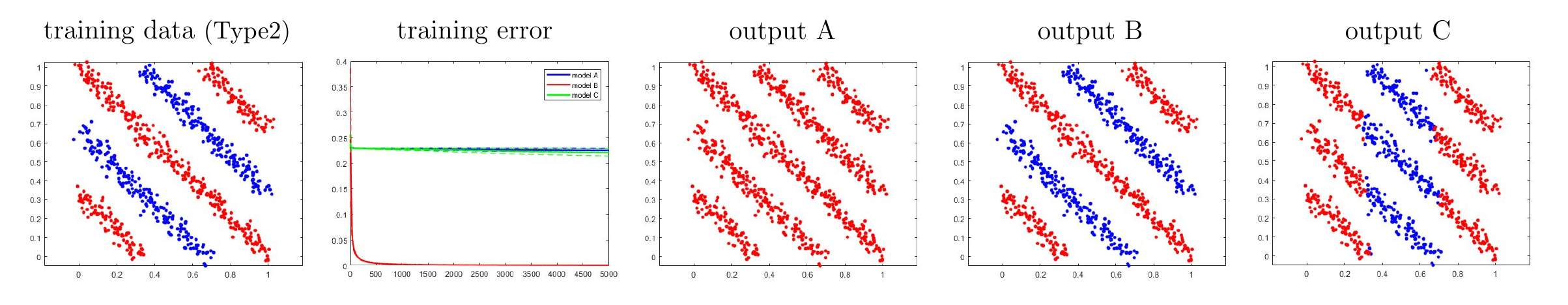}
\caption{Illustraion of the training data of `Type2'. The training error graph shows mean and standard deviation of the training errors for each model (solid line is mean and dashed line is standard deviation). The output is a training result of each model.}
\label{fig:toy2}
\end{figure*}

Next, we added the periodic information $\left(\cos\left(x_1+x_2\right)\right)$ to the model B instead of the distance information, and conducted the same experiment for the training data of `Type2'. Only model B properly classifed the training data, as shown in Fig. \ref{fig:toy2}.
In the case of `Type2' data, model C cannot obtain the information added to model B while it can in the case of `Type1' data, because of the limitation of a piecewise linear expression with a few nodes. As a result, adding information to the network had the advantage of allowing the given network to use information that cannot be obtained by learning.
From these two experiments, we observed that increasing the number of nodes in the network was more efficient in terms of the data dimension, and that adding information to the additional nodes was more beneficial than learning the relationship from the data. In the next section, we will deal with the deep learning research where our analysis was being applied.

\section{Experiments \& Results}
\begin{figure*}[!h]
		\centering
		\includegraphics[width=.945\textwidth]{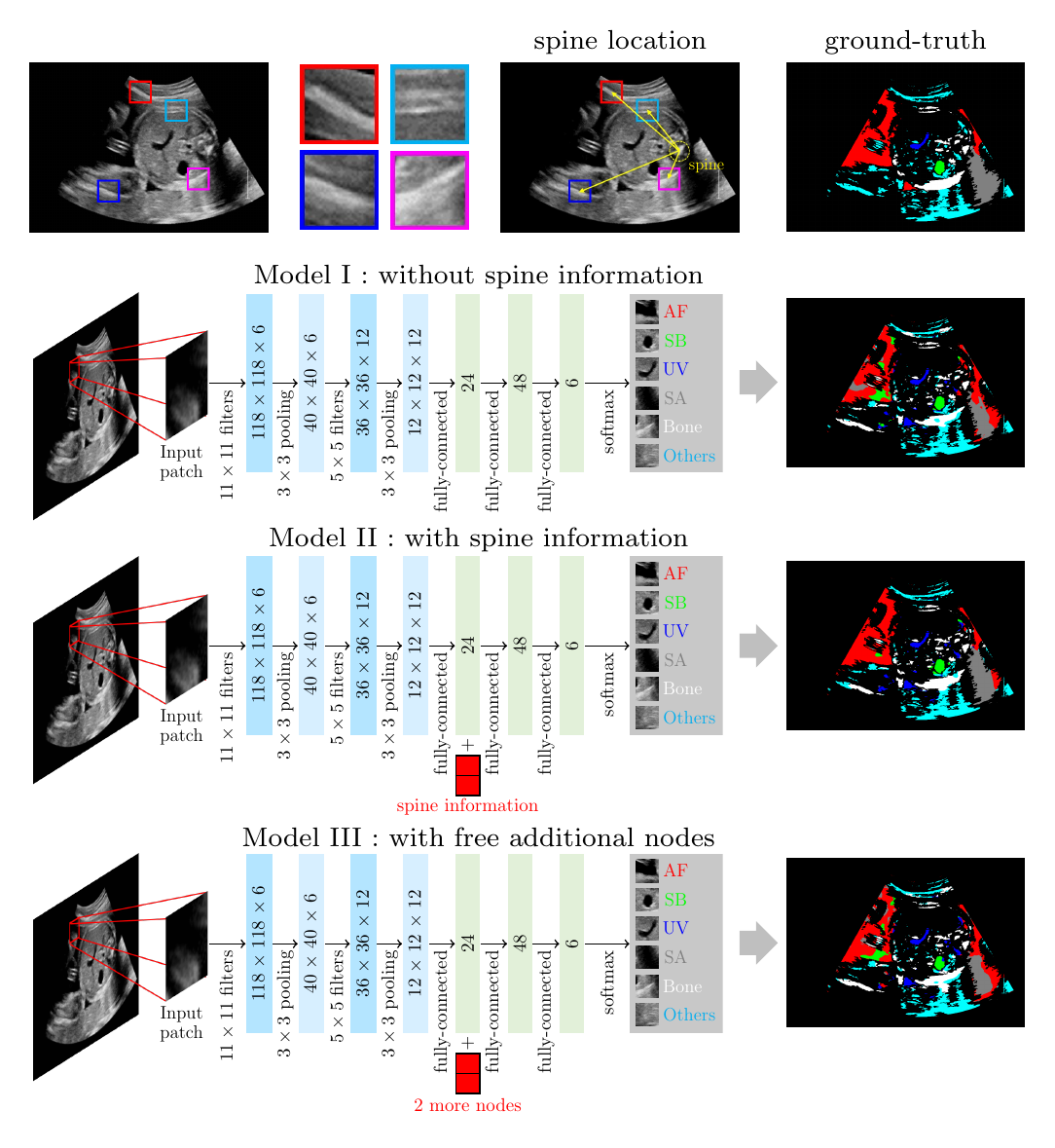}
		\caption{Comparing models with and without concatenation. }
		\label{fig:experiment_summary}	
	\end{figure*}

To support our analysis, we replicated the experiment done by Kim \etal\cite{Kim2018}.
The paper proposes an automatic fetal abdominal circumference (AC) estimation from 2D ultrasound data by using several specially designed deep neural networks that take into account clinicians' decisions, anatomical structures, and the characteristics of ultrasound images.
The proposed method has three steps: an initial AC estimation, an AC measurement, and a plane acceptance check.
These processes, require semantically segmented ultrasound images which are classified into six classes (`amniotic fluid', `fetal stomach bubble', `umbilical vein', `shadowing artifact', `bones', and `others').
The sementic segmentation is done by a neural network which uses an image patch of size $128\times 128$ to classify each pixel of the image. However, as shown in Fig. \ref{fig:experiment_summary}, the class `bone' and the class `others' are hard to classify if we rely only on image patches, because the local patterns are similar.
In order to overcome this problem, the paper used the distance in relation to the spine as a global landmark for each patch to know where they are from.

We compared the models without spine information (model I), with additional nodes of spine information (model II) and with free additional nodes (model III) which are described in Fig. \ref{fig:experiment_summary}. We used 67,894 labeled image patch training data to train the models until the 3,000th epoch. Additional details of the environment of the experiment are outlined in \cite{Kim2018}.

\begin{figure}[h]
\centering
\begin{tabular}{c}
\includegraphics[width=.45\textwidth]{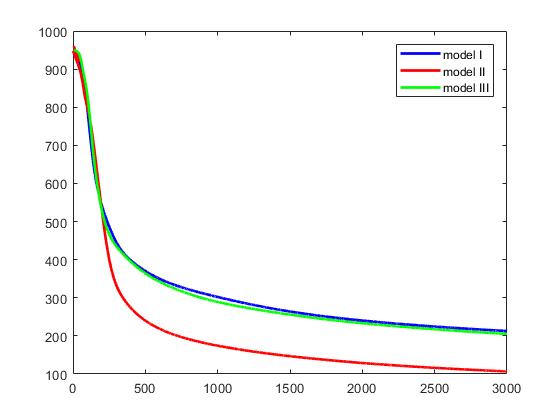}
\end{tabular}
\caption{The training error graph of each models in Fig \ref{fig:experiment_summary}.}
\label{fig:result1-2}
\end{figure}

As shown in Fig. \ref{fig:experiment_summary}, model I and III misclassify a large area, while model II achieves a better classification result. Specifically, models I and III are misclassifiying the classes `others', `bones', `shadowing artifact' and `amniotic fluids'. For these models, it is hard to distinguish between the class `bone' and the class `others'. Besides, it is hard to distinguish `shadowing artifact' and `amniotic fluid' with only local patterns. In other words, without the spine  information, it is hard to classify the patches that require the knowledge of global information.
Fig. \ref{fig:result1-2} shows the training error graph for the three models described above. Here, we can see that the model II significantly outperforms I and III.
The experiment shows that any training without adding the spine information fails to achieve an acceptable result, because the position in relation to spine is crucial in classifying every pixel of a fetal ultrasound image.

\begin{figure}[!h]
\centering
\includegraphics[width=.38\textwidth]{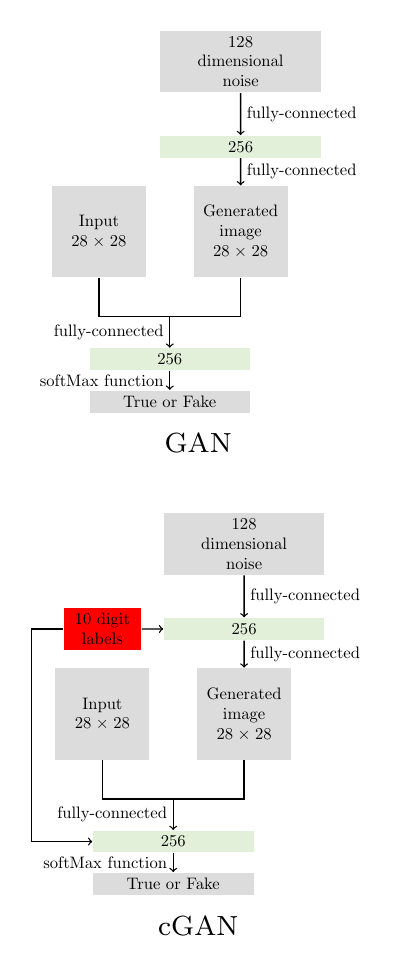}
\caption{Network architectures of the GAN and the cGAN.}
\label{fig:GanStructs}
\end{figure}

Furthermore, we compared the results between GAN and cGAN with MNIST dataset.
The GAN is an unsupervised deep learning method to generate new data from a given training dataset by using two competing neural networks: a generative model ($G$) and a discriminative model ($D$).
The GAN is trained by solving the min-max problem as follows:
\begin{eqnarray}
\min\limits_{G}\max\limits_{D}(\Bbb E_{\x\sim p_{data}(\x)}[\log D(\x)]+~~~~~~~~~~~~ \nonumber \\
\Bbb E_{\z\sim p_{\z}(\z)}[\log (1-D(G(\z)))])
\end{eqnarray}
Here, $p_{\z}(\z)$ is a prior noise distribution, $p_{data}(\x)$ is a data distribution, $\x$ is the training data sampled from the data distribution, and $\z$ is the noise sampled from $p_{\z}(\z)$ used for generative model.
Due to the limitation of GAN being unsupervised, we cannot control what kind of data GAN generates.
On contrast, the cGAN can control what kind of data it generates by using condition $\y$ as supplemantary information as described in Fig. \ref{fig:GanStructs}.
We used the label data of MNIST as $\y$ in our experiment.
The cGAN is trained by including the $\y$ in the min-max problem of GAN as follows:
\begin{eqnarray}
\min\limits_{G}\max\limits_{D}(\Bbb E_{(\x, \y)\sim p_{data}(\x,\y)}[\log D(\x|\y)]+~~~~~~~~~ \nonumber \\
\Bbb E_{\z\sim p_{\z}(\z)}[\log (1-D(G(\z|\y)))])
\label{eq:cGAN}
\end{eqnarray}

The Table \ref{tbl:cGANGAN} shows the comparison between cGAN and GAN with a Gaussian Parzen window log-likelihood estimate for the test dataset.
The cGAN has lower likelihood values than that of GAN. These values show that the cGAN generates a data distribution closer to the test data distribution compared to that of GAN.
\begin{table}[h]
   \caption{Parzen window based log-likelihood estimate}
   \centering
   \begin{tabular}{|c | c|  c|}
      \hline
& GAN & cGAN \\
      \hline
log-likelihood & $212\pm 2.1$ & $122\pm 1.9$\\
      \hline
   \end{tabular}
\label{tbl:cGANGAN}
\end{table}
\begin{figure*}[!h]
\centering
\includegraphics[width=.8\textwidth]{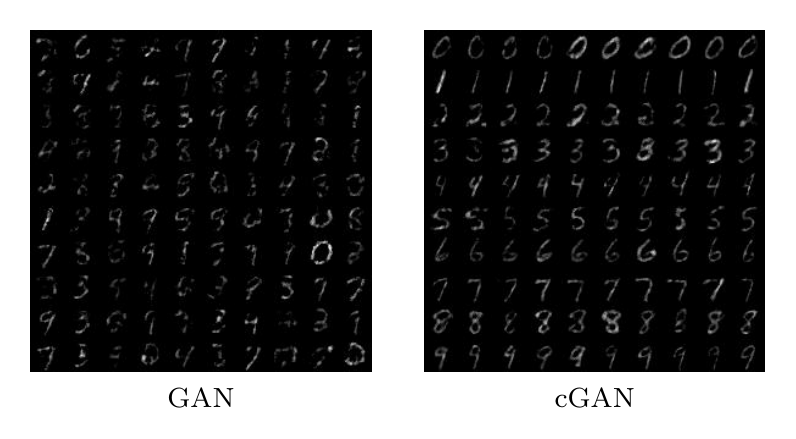}
\caption{MNIST-like images generated by GAN (left) and cGAN (right).}
\label{fig:GANcGAN}
\end{figure*}
Fig. \ref{fig:GANcGAN} shows the images generated by GAN and cGAN. The left image is generated by GAN using 100 different $\z$ sampled from a standard normal distribution of 128 dimension. The right image is generated by cGAN using the same $\z$ with condition $\y$.
Because the condition $\y$ can be controlled, we can select the mode of each generated data.
Additionally, cGAN generates more clear images than GAN in general.
This experiment shows that adding the condition as supplementary information to GAN helps the generative model to achieve a better data distribution.


As a conclusion, we confirmed that adding the nodes containing prior information improves learnability of the network by these two experiments.



\section{Discussion and Conclusion}

This paper analyzes how concatenation of additional information to a neural network affects the training efficiency of the networks. We compared and analyzed models with and without additional information using a simple toy model. We also confirmed our analysis by applying our method to segmentation of ultrasound images and by comparing cGAN and GAN.
We confirmed that adding supplementary information to a network enhances the learnability of the network by disentangling data distribution.

In deep learning networks for image analysis, patch-based methods which use patch-level images as its input data are often used, because it is much easier and more reliable to learn a model for small image patches than for the whole image, and computations are significantly reduced if they are applied on small patches and not on whole image \cite{Karimi2016}.
However, there are many cases where patch-based methods fail to give a robust result due to their nature of using only local information.
For these tasks, we expect that using a few supplementary axes with prior information will enhance their performance dramatically, similarly to our ultrasound experiment.

A deep neural network is vulnerable to adversarial attack which can give an absolutely wrong output with a small perturbation of input \cite{Fawzi2018, Finlayson2018, Goodfellow2014b}.
In our experiment, we observed that the magnitude of parameters related to the supplementary nodes are bigger compared to the magnitude of the others in trained network.
The bigger magnitude of parameters suggests the possibility of supplementary nodes influencing the output more than each pixel of the input image.
Therefore, we expect that supplementary nodes might be resistant to an adversarial attack because of their strong influence on the classification result.

Further research is necessary to expand the single layer analysis to the convolution layer and multi layer cases of deep neural networks. 
Furthermore, applying our method on various experiments would be necessary.

\section*{Acknowledgements}
This work was supported by the Samsung Science $\&$ Technology Foundation (No. SSTF-BA1402-01).

\ifCLASSOPTIONcaptionsoff
  \newpage
\fi




\begin{thebibliography}{}
\bibitem{Dorj2018}
U. O. Dorj, K. K. Lee, J. Y. Choi, and M. Lee, The skin cancer classification using deep convolutional neural network, {\it Multimedia Tools and Applications (pp. 1-16),} 2018.

\bibitem{Eigen2013}
D. Eigen, J. Rolfe, R. Fergus, and Y. LeCun, Understanding deep architectures using a recursive convolutional network, {\it arXiv preprint arXiv:1312.1847}, 2013.

\bibitem{Fawzi2018}
A. Fawzi, H. Fawzi, and O. Fawzi, Adversarial vulnerability for any classifier, {\it arXiv preprint arXiv:1802.08686.}, 2018.

\bibitem{Finlayson2018}
S. G. Finlayson, I. S. Kohane, and A. L. Beam,   Adversarial Attacks Against Medical Deep Learning Systems, {\it arXiv preprint arXiv:1804.05296.}, 2018.

\bibitem{Fraz2012}
M. M. Fraz, P. Remagnino, A. Hoppe, B. Uyyanonvara, A. R. Rudnicka, C. G. Owen, and S. A. Barman, Blood vessel segmentation methodologies in retinal images–a survey, {\it Computer methods and programs in biomedicine}, 108(1), 407-433, 2012.

\bibitem{Glorot2010}
X. Glorot and Y. Bengio, Understanding the difficulty of training deep feedforward neural networks, {\it In Proceedings of the thirteenth international conference on artificial intelligence and statistics (pp. 249-256)}, 2010.

\bibitem{Goodfellow2014}
I. Goodfellow, et al. Generative adversarial nets, {\it In: Advances in neural information processing systems (pp. 2672-2680)}, 2014.

\bibitem{Goodfellow2014b}
I. J. Goodfellow, J. Shlens, C. Szegedy, Explaining and Harnessing Adversarial Examples, {\it arXiv preprint arXiv:1412.6572.}, 2014.

\bibitem{Hyun2018}
C. M. Hyun, H. P. Kim, S. M. Lee, S. Lee, and J. K. Seo, Deep learning for undersampled MRI reconstruction, {\it Physics in medicine and biology}, 2018.

\bibitem{Karimi2016}
D. Karimi and R. K. Ward, Patch-based models and algorithms for image processing: a review of the basic principles and methods, and their application in computed tomography, {\it International journal of computer assisted radiology and surgery, 11(10)}, (pp. 1765-1777), 2016.

\bibitem{Kim2018}
B. Kim \etal, Machine-learning-based automatic identification of fetal abdominal circumference from ultrasound images, {\it Physiological Measurement. 39. 10.1088/1361-6579/aae255.}, 2018


\bibitem{LeCun2015}
Y. A. LeCun, Y. Bengio and G. Hinton, Deep learning, {\it  Nature} {\bf521 } (pp. 436-444), 2015
		

\bibitem{Mirza2014}
M. Mirza and S. Osindero, Conditional generative adversarial nets, {\it arXiv preprint arXiv:1411.1784}, 2014.

\bibitem{Pascanu2013}
R. Pascanu, G. Montufar, and Y. Bengio, On the number of response regions of deep feed forward networks with piece-wise linear activations, {\it arXiv preprint arXiv:1312.6098}, 2013.

\bibitem{Rahaman2018}
N. Rahaman, D. Arpit, A. Baratin, F. Draxler, M. Lin, F. A. Hamprecht, Y. Bengio, and A. Courville, On the Spectral Bias of Deep Neural Networks, {\it arXiv submitted arXiv:1806.08734 [stat.ML]}, 2018.



\bibitem{Yi2018}
X. Yi, E. Walia, and P. Babyn, Generative adversarial network in medical imaging: A review, {\it arXiv
preprint arXiv:1809.07294}, 2018.

\bibitem{Zeiler2014}
M. D. Zeiler and R. Fergus, Visualizing and understanding convolutional networks, {\it In European conference on computer vision (pp. 818-833),} Springer, Cham, 2014.

\bibitem{Zhang2017}
C. Zhang, S. Bengio, M. Hardt, B. Recht, and O. Vinyals. Understanding deep learning
requires rethinking generalization, {\it ICLR}, 2017.
\end{thebibliography}
%

%


\vfill


\end{document}